\documentclass[11pt,a4paper]{article}
\usepackage[hyperref]{acl2021}
\usepackage{times}
\usepackage{latexsym}

\usepackage{microtype}

\aclfinalcopy 

\newcommand\x{{\mathbf{x}}}
\newcommand\h{{\mathbf{h}}}
\newcommand\argmax{\mathrm{argmax}}
\title{Overcoming Poor Word Embeddings with Word Definitions}

\author{Christopher Malon \\
  NEC Laboratories America \\
  Princeton, NJ 08540 \\
  \texttt{malon@nec-labs.com} \\}

\date{}

\begin{document}
\maketitle
\begin{abstract}
Modern natural language understanding models depend on pretrained
subword embeddings, but applications may need to reason about words
that were never or rarely seen during pretraining.
We show that examples that depend critically on a rarer word 
are more challenging for natural language inference models.
Then we explore how a model could learn to use definitions,
provided in natural text, to overcome this handicap.
Our model's understanding of a definition is usually weaker than a
well-modeled word embedding, but it recovers most of the performance gap
from using a completely untrained word.
\end{abstract}

\section{Introduction}

The reliance of natural language understanding models
on the information in pre-trained word embeddings limits
these models from being applied reliably to rare words or technical vocabulary.
To overcome this vulnerability, a model must be able to compensate for a
poorly modeled word embedding with background knowledge to complete the
required task.

For example, a natural language inference (NLI) model based
on pre-2020 word embeddings may not be able to deduce
from ``Jack has COVID'' that ``Jack is sick.''
By providing the definition, ``COVID is a respiratory disease,'' we want
to assist this classification.

We describe a general procedure for enhancing a classification
model such as natural language inference (NLI) or sentiment classification,
to perform the same task on sequences including poorly modeled words
using definitions of those words.
From the training set $\mathcal{T}$ of the original model,
we construct an augmented training set $\mathcal{T}^\prime$
for a model that may accept the same token sequence optionally concatenated
with a word definition.  In the case of NLI,
where there are two token sequences, the definition is concatenated to the
premise sequence.  Because $\mathcal{T}^\prime$ has the same form
as $\mathcal{T}$, a model accepting the augmented information may be
trained in the same way as the original model.

Because there are not enough truly untrained words like ``COVID'' in
natural examples, we probe performance by scrambling real words so that
their word embedding becomes useless, and supplying definitions.
Our method recovers most of the performance lost by scrambling.
Moreover, the proposed technique removes biases in more {\em ad hoc} solutions
like adding definitions to examples without special training.

\section{Related Work}

We focus on NLI because it depends
more deeply on word meaning than sentiment or topic classification tasks.
\citet{chen-etal-2018-neural-natural} pioneered the addition of
background information to an NLI model's classification on a per-example
basis, augmenting a sequence of token embeddings with features encoding
WordNet relations between pairs of words, to achieve a 0.6\% improvement
on the SNLI \citep{bowman-etal-2015-large} task.
Besides this explicit reasoning approach,
implicit reasoning over background knowledge can be achieved if one updates
the base model itself with background information.
\citet{lauscher-etal-2020-common} follows this approach to add information from
ConceptNet \citep{conceptnet} and the Open Mind Common Sense corpus
\citep{omcs}
through a fine-tuned adapter added to a pretrained language model,
achieving better performance on subsets of NLI examples that are known
to require world knowledge.
\citet{leapthought} explore the interplay between explicitly added knowledge
and implicitly stored knowledge on artificially constructed NLI problems
that require counting or relations from a taxonomy.
 
In the above works, explicit background information comes from
a taxonomy or knowledge base.  Only a few studies have worked with
definition text directly, and not in the context of NLI.
\citet{tissier-etal-2017-dict2vec} used definitions to create embeddings
for better performance on word similarity tasks,
compared to word2vec \citep{mikolov2013} and fastText
\citep{bojanowski2017} while maintaining performance on text classification.
Recently, \citet{debiasing} used definitions to remove biases from pretrained
word embeddings while maintaining coreference resolution accuracy.
In contrast, our work reasons with natural language
definitions without forming a new embedding.

\section{Methods}

\subsection{Critical words}

The enhanced training set $\mathcal{T}^\prime$ will be built by providing
definitions for words in existing examples, while obfuscating the existing
embeddings of those words.
If a random word of the original text is obfuscated, the classification still
may be determined or strongly biased by the remaining words.
To ensure the definitions matter, we select carefully.

To explain which words of a text are important for classification,
\citet{kim-etal-2020-interpretation} introduced the idea of input
marginalization.  Given a sequence of tokens $\x$, such that $\x_{-i}$
represents the sequence without the $i$th token $x_i$, they marginalize
the probability of predicting a class $y_c$ over possible
replacement words $\tilde{x_i}$ in the vocabulary $\mathcal{V}$ as
\begin{equation}
p(y_c | \x_{-i}) = \sum_{\tilde{x_i} \in \mathcal{V}}
p(y_c | \tilde{x_i}, \x_{-i}) p(\tilde{x_i} | \x_{-i})
\end{equation}
and then compare $p(y_c | \x_{-i})$ to $p(y_c | \x)$ to quantify the
importance of $x_i$.  The probabilities $p(\tilde{x_i} | \x_{-i})$ are computed
by a language model.

We simplify by looking only at the classification and not the probability.
Like \citet{kim-etal-2020-interpretation}, we truncate the computation
of $p(y_c | \tilde{x_i}, \x_{-i})$ to words such that
$p(\tilde{x_i} | \x_{-i})$ exceeds a threshold, here .05.
Ultimately we mark a word $x_i$
as a {\em critical word} if there exists a replacement $\tilde{x_i}$
such that 
\begin{equation}
\argmax_y p(y | \tilde{x_i}, \x_{-i}) \neq \argmax_y p(y | \x)
\end{equation}
and
\begin{equation}
p(\tilde{x_i} | x_{-i}) > .05  \ldotp
\end{equation}
Additionally we require that the word not appear more than once in the
example, because the meaning of repeated words usually impacts the
classification less than the fact that they all match.
Table~\ref{tab:critical} shows an example.

\begin{table}[htb]
\begin{tabular}{p{.75in}p{2in}}
\hline
Premise & A young man sits, looking out of a {\em train} [side
$\rightarrow$ Neutral, small $\rightarrow$ Neutral]  window. \\
Hypothesis & The man is in his room. \\
Label & Contradiction \\
\hline
\end{tabular}
\caption{An SNLI example, with critical words shown in italics and
replacements shown in brackets.} \label{tab:critical}
\end{table}

A technicality remains because our classification models use subwords
as tokens, whereas we consider replacements of whole words
returned by \verb+pattern.en+.  We remove all subwords of $x_i$
when forming $\x_{-i}$, but we consider only replacements $\tilde{x_i}$
that are a single subword long.

\subsection{Definitions}

We use Wiktionary as a source of definitions.  The code of
\citet{tissier-etal-2017-dict2vec} downloaded definitions from four
commercial online dictionaries, but these are no longer freely available
online as of January 2021.  When possible, we
look for a definition in the Simple English Wiktionary, because these
definitions refer to more common usages of words and are written using
simpler language.  If one is not found, we consult the regular English
Wiktionary.\footnote{We use the 2018-02-01 dumps.}

To define a word, first we find its part of speech in the original context
and lemmatize the word using the \verb+pattern.en+ library \cite{pattern-en}.
Then we look for a section labeled ``English'' in the retrieved Wiktionary
article, and for a subsection for the part of speech we identified.
We extract the first numbered definition in this subsection.

There is no guarantee that this sense of the word matches the sense
used in the text, but since the word embedding for any other word would
be determined only by its spelling, we expect good performance even if a
different sense of the word is chosen.  In practice, we find that this
method usually gives us short, simple definitions that match the usage
in the original text.

When defining a word, we always write its definition as ``{\em word} means:
{\em definition}.''  This common format ensures that the definitions
and the word being defined can be recognized easily by the classifier.

\subsection{Enhancing a model}

\subsubsection{Without scrambling}

Consider an example $(\x, y_c) \in \mathcal{T}$.  If the example has a critical
word $x_i \in \x$ that appears only once in the example, and
$\tilde{x_i}$ is the most likely replacement word that changes
the classification, we let $\x^\prime$ denote the sequence where $x_i$
is replaced by $\tilde{x_i}$, and let
$y^\prime_c = \argmax_y p(y | \x^\prime)$.
If definitions $\h_i$ and $\h^\prime_i$
for $x_i$ and $\tilde{x_i}$ are found by the method described above, 
we add $(\x, \h_i, y_c)$ and $(\x^\prime, \h^\prime_i, y^\prime_c)$
to the enhanced training set $\mathcal{T}^\prime$.

\subsubsection{With scrambling}

Scrambling a word prevents the model from relying on a useful word embedding.
In this protocol, we generate random strings of letters, of random length
between four and twelve letters, to substitute for $x_i$ and $\tilde{x_i}$,
while still using the definitions of the the original words.
If the original words appear in their own
definitions, those occurrences are also replaced by the same strings.
Unfortunately, the random strings lose any morphological features
of the original words.

Table~\ref{tab:background} shows an NLI example and the corresponding
examples generated for the enhanced training set.

\begin{table}[htb]
\begin{tabular}{p{.75in}p{2in}}
\hline
Original & A blond man is drinking from a public fountain. / The man is
drinking water. / Entailment \\
Scrambled word & a blond man is drinking from a public yfcqudqqg.
yfcqudqqg means: a natural source of water; a spring. /
the man is drinking water. / Entailment \\
Scrambled alternate & a blond man is drinking from a public lxuehdeig.
lxuehdeig means: lxuehdeig is a transparent solid and is usually clear.
windows and eyeglasses are made from it, as well as drinking glasses. /
the man is drinking water. / Neutral \\
\hline
\end{tabular}
\caption{Adding background information to examples from SNLI}
\label{tab:background}
\end{table}

\section{Experiments}

\subsection{Setup}

We consider the SNLI task \citep{bowman-etal-2015-large}.
We fine-tune an XLNet model \citep{xlnet}, because it achieves near
state-of-the-art performance on SNLI and outperforms Roberta \citep{roberta}
and BERT \citep{bert}
on later rounds of adversarial annotation for ANLI
\citep{nie-etal-2020-adversarial}.
Due to computing constraints we use the base, cased
model.  Training is run for three epochs distributed across 4 GPU's, with a
batch size of 10 on each, a learning rate of $5 \times 10^{-5}$, 120
warmup steps, a single gradient accumulation step, and a maximum sequence
length of 384.

For the language model probabilities $p(\tilde{x_i} | \x_{-i})$,
pretrained BERT (base, uncased) is used rather than XLNet because the XLNet
probabilities have been observed to be very noisy on short
sequences.\footnote{https://github.com/huggingface/transformers/issues/4343}

One test set $SNLI_{crit}^{full}$ is constructed in the same way as
the augmented training set, but our main test set
$SNLI_{crit}^{true}$ is additionally constrained to use only examples
of the form
$(\x, \h_i, y_c)$ where $y_c$ is the original label, because labels for
the examples $(\x^\prime, \h^\prime_i, y^\prime_c)$ might be incorrect.

Not every SNLI example has a critical word, and we do not always find a
definition with the right part of speech in Wiktionary.
Our training and test sets have 272,492 and 2,457 examples
({\em vs}. 549,367 and 9,824 in SNLI).  
All of our derived datasets are available for
download.\footnote{https://figshare.com/s/edd5dc26b78817098b72}

\subsection{Results}

Table~\ref{tab:results} compares various training protocols.

\begin{table}[htb]
\begin{tabular}{p{2in}p{.75in}}
\hline
{\bf Protocol} & $SNLI_{crit}^{true}$ \\
\hline
Original & 85.1\% \\
No scrambling, no defs & 84.6\% \\ 
No scrambling, defs & 85.2\% \\ 
Scrambling, no defs & 36.9\% \\ 
Scrambling, defs & 81.2\% \\ 
Scrambling, subs & 84.7\% \\ 
Train on normal SNLI, test on scrambled no defs & 54.1\% \\
Train on normal SNLI, test on scrambled defs & 63.8\% \\
Train on unscrambled defs, test on scrambled defs & 51.4\% \\
\hline
\end{tabular}
\caption{Comparing enhancement protocols}
\label{tab:results}
\end{table}

{\bf Our task cannot be solved well without reading definitions.}
When words are scrambled but no definitions are provided,
an SNLI model without special training achieves 54.1\%
on $SNLI_{crit}^{true}$.  If trained on $\mathcal{T}^\prime$ with
scrambled words but no definitions, a model achieves 36.9\%, which is
even lower, reflecting that the training set is constructed to prevent
a model from utilizing the contextual bias.

{\bf With definitions and scrambled words, performance is slightly below
that of using the original words.}
Our method using definitions applied to the scrambled words yields 81.2\%,
compared to 84.6\% if words are left unscrambled but no definitions are
provided.  Most of the accuracy lost by obfuscating the words is recovered,
but evidently there is slightly
more information accessible in the original word embeddings.

{\bf If alternatives to the critical words are not included,
the classifier learns biases that do not depend on the definition.}
We explore restricting the training set to verified examples
$\mathcal{T}^\prime_{true} \subset \mathcal{T}^\prime$
in the same way as the $SNLI_{crit}^{true}$, still scrambling the
critical or replaced words in the training and testing sets.
Using this subset, a model that is not given the definitions can be trained
to achieve 69.9\% performance on $SNLI_{crit}^{true}$, showing a heavy
contextual bias.  A model trained on this subset that uses the definitions
achieves marginally higher performance (82.3\%)
than the one trained on all of $\mathcal{T}^\prime$.
On the other hand, testing on $SNLI_{crit}^{full}$ yields only 72.3\%
compared to 80.3\% using the full $\mathcal{T}^\prime$,
showing that the classifier is less sensitive to the definition.

{\bf Noisy labels from replacements do not hurt accuracy much.}
The only difference between the ``original'' training protocol and
``no scrambling, no defs'' is that the original trains on $\mathcal{T}$ and
does not include examples
with replaced words and unverified labels.  Training including the
replacements reduces accuracy by 0.5\% on $SNLI_{crit}^{true}$,
which includes only verified labels.  For comparison, training and testing
on all of SNLI with the original protocol
achieves 90.4\%, so a much larger effect on accuracy must be due to
selecting harder examples for $SNLI_{crit}^{true}$.

{\bf Definitions are not well utilized without special training.}
The original SNLI model, if provided definitions
of scrambled words at test time as part of the premise,
achieves only 63.8\%, compared to 81.2\% for our specially trained model.

{\bf If the defined words are not scrambled, the classifier uses the
original embedding and ignores the definitions.}
Training with definitions but no scrambling, 85.2\% accuracy is
achieved, but this trained model is unable to use the definitions when
words are scrambled: it achieves 51.4\% on that test set.

{\bf We have not discovered a way to combine the benefit of the definitions
with the knowledge in the original word embedding.}
To force the model to use both techniques,
we prepare a version of the training set which is half scrambled and
half unscrambled.  This model achieves 83.5\% on the unscrambled test
set, below the result if no definitions are provided.

{\bf Definitions are not simply being memorized.}
We selected the subset $SNLI_{crit}^{new}$ of $SNLI_{crit}^{true}$ consisting
of the 44 examples in which the defined word was not defined in a
training example.  The definition scrambled model achieves 68.2\%
on this set, well above 45.5\% for the original SNLI model reading the
scrambled words and definitions but without special training.
Remembering a definition from training is thus an advantage (reflected
in the higher 81.2\% accuracy on $SNLI_{crit}^{true}$), but not the whole
capability.

{\bf Definition reasoning is harder than simple substitutions.}
When definitions are given as one-word substitutions,
in the form ``{\em scrambled} means: {\em original}''
instead of ``{\em scrambled} means: {\em definition}'',
the model achieves 84.7\% on $SNLI_{crit}^{true}$ compared
to 81.2\% using the definition text.  Of course this is not a possibility
for rare words that are not synonyms of a word that has been well trained,
but it suggests
that the kind of multi-hop reasoning in which words just have to be
matched in sequence is easier than understanding a text definition.

\subsection{A hard subset of SNLI}

By construction of the SentencePiece dictionary
\citep{kudo-richardson-2018-sentencepiece}, only the most frequent words
in the training data of the XLNet language model are represented as
single tokens.  Other words are tokenized by multiple subwords.
Sometimes the subwords
reflect a morphological change to a well-modeled word, such as a change
in tense or plurality.  The language model probably understands these
changes well and the subwords give important hints.  The lemma form of
a word strips many morphological features, so when the lemma
form of a word has multiple subwords, the basic concept may be less
frequently encountered in training.
We hypothesize that such words are less well understood by
the language model.

To test this hypothesis, we construct a subset $SNLI_{multi}^{true}$
of the test set, consisting of examples where a critical word exists
whose lemma form spans multiple subwords, and for which an appropriate
definition can be found in Wiktionary.  This set consists of 332 test
examples.  The critical word used may be different from the one
chosen for $SNLI_{crit}^{true}$.  This subset is indeed harder:
the XLNet model trained on
all of SNLI attains only 77.7\% on this subset using no definitions,
compared to 90.4\% on the original test set.

In Table~\ref{tab:hard} we apply various models constructed in the
previous subsection to this hard test set.
Ideally, a model leveraging definitions could compensate for these
weaker word embeddings, but the method here does not do so.

\begin{table}
\begin{centering}
\begin{tabular}{p{2in}p{.75in}}
\hline
{\bf Protocol} & $SNLI_{multi}^{true}$ \\
\hline
Normal SNLI on unscrambled & 77.7\% \\
Defs \& unscrambled on defs \& unscrambled & 77.1\% \\
Defs \& some scrambling on defs \& unscrambled & 73.8\% \\
Defs \& scrambled on defs \& scrambled & 69.9\% \\
Defs \& scrambled on defs \& unscrambled & 62.7\% \\
\hline
\end{tabular}
\caption{Performance on the hard SNLI subset}
\label{tab:hard}
\end{centering}
\end{table}

\section{Conclusion}

This work shows how a model's training may be enhanced to support
reasoning with definitions in natural text, to handle cases where word
embeddings are not useful.  Our method forces the definitions to be
considered and avoids the application of biases independent of the definition.
Using the approach, entailment examples like ``Jack has COVID / Jack is sick''
that are misclassified by an XLNet trained on normal SNLI are
correctly recognized as entailment when a definition ``COVID is a
respiratory disease'' is added.
Methods that can leverage definitions without losing the advantage of
partially useful word embeddings are still needed.
In an application, it also will be necessary to
select the words that would benefit from definitions, and to make a
model that can accept multiple definitions.

\bibliographystyle{acl_natbib}
\bibliography{anthology,acl2021-arxiv}

\begin{thebibliography}{17}
\expandafter\ifx\csname natexlab\endcsname\relax\def\natexlab#1{#1}\fi

\bibitem[{Bojanowski et~al.(2017)Bojanowski, Grave, Joulin, and
  Mikolov}]{bojanowski2017}
Piotr Bojanowski, Edouard Grave, Armand Joulin, and Tomas Mikolov. 2017.
\newblock Enriching word vectors with subword information.
\newblock \emph{arXiv preprint}, 1607.04606.

\bibitem[{Bowman et~al.(2015)Bowman, Angeli, Potts, and
  Manning}]{bowman-etal-2015-large}
Samuel~R. Bowman, Gabor Angeli, Christopher Potts, and Christopher~D. Manning.
  2015.
\newblock A large annotated corpus for learning natural language inference.
\newblock In \emph{Proceedings of the 2015 Conference on Empirical Methods in
  Natural Language Processing}, pages 632--642, Lisbon, Portugal. Association
  for Computational Linguistics.

\bibitem[{Chen et~al.(2018)Chen, Zhu, Ling, Inkpen, and
  Wei}]{chen-etal-2018-neural-natural}
Qian Chen, Xiaodan Zhu, Zhen-Hua Ling, Diana Inkpen, and Si~Wei. 2018.
\newblock Neural natural language inference models enhanced with external knowledge.
\newblock In \emph{Proceedings of the 56th Annual Meeting of the Association
  for Computational Linguistics (Volume 1: Long Papers)}, pages 2406--2417,
  Melbourne, Australia. Association for Computational Linguistics.

\bibitem[{Devlin et~al.(2019)Devlin, Chang, Lee, and Toutanova}]{bert}
Jacob Devlin, Ming-Wei Chang, Kenton Lee, and Kristina Toutanova. 2019.
\newblock Bert: Pre-training of deep bidirectional transformers for language
  understanding.
\newblock \emph{arXiv preprint}, 1810.04805.

\bibitem[{Kaneko and Bollegala(2021)}]{debiasing}
Masahiro Kaneko and Danushka Bollegala. 2021.
\newblock Dictionary-based debiasing of pre-trained word embeddings.
\newblock \emph{arXiv preprint}, 2101.09525.

\bibitem[{Kim et~al.(2020)Kim, Yi, Kim, and
  Yoon}]{kim-etal-2020-interpretation}
Siwon Kim, Jihun Yi, Eunji Kim, and Sungroh Yoon. 2020.
\newblock Interpretation of {NLP} models through input marginalization.
\newblock In \emph{Proceedings of the 2020 Conference on Empirical Methods in
  Natural Language Processing (EMNLP)}, pages 3154--3167, Online. Association
  for Computational Linguistics.

\bibitem[{Kudo and Richardson(2018)}]{kudo-richardson-2018-sentencepiece}
Taku Kudo and John Richardson. 2018.
\newblock {S}entence{P}iece: A simple and language independent subword tokenizer and detokenizer for neural text processing.
\newblock In \emph{Proceedings of the 2018 Conference on Empirical Methods in
  Natural Language Processing: System Demonstrations}, pages 66--71, Brussels,
  Belgium. Association for Computational Linguistics.

\bibitem[{Lauscher et~al.(2020)Lauscher, Majewska, Ribeiro, Gurevych, Rozanov,
  and Glava{\v{s}}}]{lauscher-etal-2020-common}
Anne Lauscher, Olga Majewska, Leonardo F.~R. Ribeiro, Iryna Gurevych, Nikolai
  Rozanov, and Goran Glava{\v{s}}. 2020.
\newblock Common sense or world knowledge? investigating adapter-based knowledge injection into pretrained transformers.
\newblock In \emph{Proceedings of Deep Learning Inside Out (DeeLIO): The First
  Workshop on Knowledge Extraction and Integration for Deep Learning
  Architectures}, pages 43--49, Online. Association for Computational
  Linguistics.

\bibitem[{Liu et~al.(2019)Liu, Ott, Goyal, Du, Joshi, Chen, Levy, Lewis,
  Zettlemoyer, and Stoyanov}]{roberta}
Yinhan Liu, Myle Ott, Naman Goyal, Jingfei Du, Mandar Joshi, Danqi Chen, Omer
  Levy, Mike Lewis, Luke Zettlemoyer, and Veselin Stoyanov. 2019.
\newblock Roberta: A robustly optimized bert pretraining approach.
\newblock \emph{arXiv preprint}, 1907.11692.

\bibitem[{Mikolov et~al.(2013)Mikolov, Chen, Corrado, and Dean}]{mikolov2013}
Tomas Mikolov, Kai Chen, Greg Corrado, and Jeffrey Dean. 2013.
\newblock Efficient estimation of word representations in vector space.
\newblock \emph{arXiv preprint}, 1301.3781.

\bibitem[{Nie et~al.(2020)Nie, Williams, Dinan, Bansal, Weston, and
  Kiela}]{nie-etal-2020-adversarial}
Yixin Nie, Adina Williams, Emily Dinan, Mohit Bansal, Jason Weston, and Douwe
  Kiela. 2020.
\newblock Adversarial {NLI}: A new benchmark for natural language understanding.
\newblock In \emph{Proceedings of the 58th Annual Meeting of the Association
  for Computational Linguistics}, pages 4885--4901, Online. Association for
  Computational Linguistics.

\bibitem[{Singh et~al.(2002)Singh, Lin, Mueller, Lim, Perkins, and
  Li~Zhu}]{omcs}
Push Singh, Thomas Lin, Erik~T. Mueller, Grace Lim, Travell Perkins, and Wan
  Li~Zhu. 2002.
\newblock Open mind common sense: Knowledge acquisition from the general
  public.
\newblock In \emph{On the Move to Meaningful Internet Systems 2002: CoopIS,
  DOA, and ODBASE}, pages 1223--1237, Berlin, Heidelberg. Springer Berlin
  Heidelberg.

\bibitem[{Smedt and Daelemans(2012)}]{pattern-en}
Tom~De Smedt and Walter Daelemans. 2012.
\newblock Pattern for python.
\newblock \emph{Journal of Machine Learning Research}, 13(66):2063--2067.

\bibitem[{Speer et~al.(2018)Speer, Chin, and Havasi}]{conceptnet}
Robyn Speer, Joshua Chin, and Catherine Havasi. 2018.
\newblock Conceptnet 5.5: An open multilingual graph of general knowledge.
\newblock \emph{arXiv preprint}, 1612.03975.

\bibitem[{Talmor et~al.(2020)Talmor, Tafjord, Clark, Goldberg, and
  Berant}]{leapthought}
Alon Talmor, Oyvind Tafjord, Peter Clark, Yoav Goldberg, and Jonathan Berant.
  2020.
\newblock Leap-of-thought: Teaching pre-trained models to systematically reason over
  implicit knowledge.
\newblock In \emph{Advances in Neural Information Processing Systems 33: Annual
  Conference on Neural Information Processing Systems 2020, NeurIPS 2020,
  December 6-12, 2020, virtual}.

\bibitem[{Tissier et~al.(2017)Tissier, Gravier, and
  Habrard}]{tissier-etal-2017-dict2vec}
Julien Tissier, Christophe Gravier, and Amaury Habrard. 2017.
\newblock {D}ict2vec : Learning word embeddings using lexical dictionaries.
\newblock In \emph{Proceedings of the 2017 Conference on Empirical Methods in
  Natural Language Processing}, pages 254--263, Copenhagen, Denmark.
  Association for Computational Linguistics.

\bibitem[{Yang et~al.(2019)Yang, Dai, Yang, Carbonell, Salakhutdinov, and
  Le}]{xlnet}
Zhilin Yang, Zihang Dai, Yiming Yang, Jaime Carbonell, Russ~R Salakhutdinov,
  and Quoc~V Le. 2019.
\newblock Xlnet: Generalized autoregressive pretraining for language understanding.
\newblock In \emph{Advances in Neural Information Processing Systems},
  volume~32, pages 5753--5763. Curran Associates, Inc.

\end{thebibliography}


\end{document}